\documentclass[letterpaper, 10 pt, conference]{ieeeconf}
\IEEEoverridecommandlockouts
\usepackage{xcolor, cite, amsmath, amssymb, xspace, indentfirst, xr, bm, verbatim, dsfont, booktabs, graphicx, url, float}
\usepackage{subfigure}
\usepackage[lined,ruled,linesnumbered]{algorithm2e}

\title{\LARGE \bf
A Novel GPU-based Parallel Implementation Scheme and\\
Performance Analysis of Robot Forward Dynamics Algorithms
}

\author{Yajue Yang, Yuanqing Wu and Jia Pan
\thanks{Yajue Yang and Jia Pan are with the Department of Mechanical and Biomedical Engineering, City University of Hong Kong, Hong Kong.}%
\thanks{Yuanqing Wu is with the Department of Industrial Engineering (DIN) University of Bologna, Italy.}%
}

\DeclareMathOperator{\ad}{ad}

\newcommand\se[1][3]{\ensuremath{se(#1)}\xspace}

\newcommand\Ji[1][]{\ensuremath{J^{-1}\ifx\\#1\\\else_{#1}\fi}\xspace}
\newcommand\SE[1][3]{\ensuremath{\mathrm{SE}(#1)}\xspace}
\newcommand\ds\displaystyle
\begin{document}

\maketitle
\thispagestyle{empty}
\pagestyle{empty}

\begin{abstract}

We propose a novel unifying scheme for parallel implementation of articulated robot dynamics algorithms. It is based on a unified Lie group notation for deriving the equations of motion of articulated robots, where various well-known forward algorithms differ only by their joint inertia matrix inversion strategies. This new scheme leads to a unified abstraction of state-of-the-art forward dynamics algorithms into combinations of block bi-diagonal and/or block tri-diagonal systems, which may be efficiently solved by parallel all-prefix-sum operations (scan) and parallel odd-even elimination (OEE) respectively. We implement the proposed scheme on a Nvidia CUDA GPU platform for the comparative study of three algorithms, namely the hybrid articulated-body inertia algorithm (ABIA), the parallel joint space inertia inversion algorithm (JSIIA) and the constrained force algorithm (CFA), and the performances are analyzed.

\end{abstract}


\externaldocument{related}
\externaldocument{oee_gpu}
\externaldocument{experiment}
\section{Introduction}

In recent years, the robotics community has witnessed increasingly wider applications of CPU-GPU heterogeneous computation for its remarkable number crunching and also task/data parallelism capabilities~\cite{nickolls2010gpu}. 
In particular, the task of solving substantial groups of inverse/forward dynamics problems for articulated robots and humanoid robots with a moderate number of links~\cite{lengagne2013generation, chretien2016gpu, todorov2012mujoco, laflin2016enhancing, fijany2013new} may now be completely run on GPU platforms instead of traditional multi-processor systems. This, in turn, opens up new research on the classic inverse/forward dynamics algorithms.
It should be emphasized that the aim of such research is not to develop essentially new algorithms under the new GPU setting, but to determine the compatibility with the GPU platform of state-of-the-art robot dynamics algorithms~\cite{featherstone1983calculation,rodriguez1991spatial,fijany1992parallel,featherstone1999divide,ploen1999coordinate}, some of which are already adapted to parallelized multi-processor systems.

In our recent study~\cite{YWP16a}, we have identified two major problems when developing and comparing robot forward dynamics algorithms on the GPU platform. First, different algorithms often come with different mathematical and notation conventions, which makes their adaptation to the GPU platform laborious and error prone. Therefore, it is preferable to select a concise and intuitive notation, to which all forward dynamics algorithms can be easily translated. Second, the earlier parallel adaptation of robot forward dynamics algorithms often relies on ad hoc parallel implementation, which makes it difficult to conduct an efficient performance comparison between different algorithms. Therefore, it is preferable to have a consistent implementation scheme of different algorithms using some well-developed generic building blocks to avoid reinventing the wheels~\cite{negrut2012leveraging}.

In this paper, we propose a novel GPU-based parallel implementation scheme and performance analysis of robot forward dynamics algorithms. First, we advocate for the use of a de facto coordinate-free Lie group notation for robot dynamics algorithms~\cite{ploen1999coordinate}. The process of translating different forward dynamics algorithms to this notation turns out to be enlightening rather than cumbersome. Second, we focus on the parallel implementation of two building blocks, namely, block bi-diagonal and block tri-diagonal systems, which are common to various robot inverse and forward dynamics algorithms. The former may be efficiently solved by the all-prefix-sum (scan) operation and the latter may be efficiently solved by the odd-even elimination (OEE) algorithm. Using the proposed scheme, we compare the running performance of three representative robot forward dynamics algorithms on a Nvidia CUDA GPU platform: the hybrid articulated-body inertia algorithm (ABIA) based on Featherstone's pioneering work~\cite{featherstone1983calculation}, the parallel joint space inertia inversion algorithm (JSIIA), and the constrained force algorithm (CFA) proposed by Fijany~\cite{fijany1992parallel}.

\subsection{Brief review of robot forward dynamic algorithms}

Fast computation of robot motion trajectories with given state and inputs is of vital importance to not only motion simulation of robotic systems with a large number of bodies~\cite{redon2005adaptive}, but also online trajectory planning~\cite{yamane2003dynamics, lee2005newton, chitchian2012particle, lengagne2013generation, chretien2016gpu} and model-based control optimization~\cite{todorov2012mujoco, erez2012trajectory}. The main challenge to efficiently solving such \emph{forward dynamics} (FD) problem originates in the recursive propagation of motions and forces along serial kinematic chains and inverting joint space inertia (JSI) matrix with a high dimension. Parallel forward dynamics algorithms have long been developed and implemented on multi-processor systems to meet such demands. The CFA proposed by Fijany \emph{et al.} is claimed to be the first truly parallel FD algorithm that has $O(\log n)$ time complexity on $O(n)$ processors~\cite{fijany1995parallel} ($n$ being the number of links in the manipulator). Featherstone later presented a recursive, divide-and-conquer algorithm (DCA) based on the articulated body inertia (ABI) theory~\cite{featherstone1999divide}. These are often compared with sequential-parallel hybrid $O(n)$ algorithms such as by Anderson and Duan~\cite{anderson1999hybrid}.

\subsection{Related work}
The following are some highly related work to our paper. \cite{chretien2016gpu, zhang1998nonrecursive} have utilized the all-prefix-sums (scan) operation to parallelize the propagation of rotation matrices when computing articulated robot inverse dynamics on GPU. Part of our contribution in this paper is to systematize the use of scan operations. 
Yamane \emph{et al.} proposed a comparative study of various robot forward dynamics algorithms, which focuses on comparing the representation of joint geometry and applicability to robots with branches and loops~\cite{yamane2003dynamics}. Our work focus instead on the building blocks of different algorithms and the different variable elimination strategies that lead to different factorization of the inverse of the joint space inertia.

\subsection{Organization of the paper}
This paper is organized as follows. In section~\ref{sec:algo_recap}, we first briefly review the recursive formulation of robot dynamics using the matrix Lie group notation. Observing that some inverse/forward dynamics algorithms derived from this formulation contain a set of bi-diagonal systems, we then introduce the parallel scan operation to accelerate these algorithms. Since not all equations can be transformed into the scan operation, section~\ref{sec:cfa} turns to exploring the CFA, a fully parallel forward dynamics algorithm, which involves solving a tri-diagonal system. Section~\ref{sec:oee-algo} then illustrates the tailored parallel OEE algorithm for solving the special tri-diagonal system in the CFA. In section~\ref{sec:experiment}, we implement the hybrid/parallel forward dynamics algorithms on a CPU/GPU platform and conduct several experiments to test their performances in different applications.
\externaldocument{oee_gpu}
\section{Robot Dynamics Algorithms Involving Block Bi-Diagonal Systems}
\label{sec:algo_recap}
Throughout this paper, we shall use the Lie group notation~\cite{murray1994mathematical,park1995lie} and follow the global matrix representation in~\cite{ploen1999coordinate} for the equations of motion of rigid bodies and articulated robots. 
We use \SE and \se to denote the special Euclidean group and its Lie algebra, and $\se^*$ to denote the dual of \se. Both \se and $\se^*$ are identified with the $6$-D vector space $\mathds R^6$ in a natural way.

\subsection{Recursive formulation of robot dynamics}

Recall that the Newton-Euler equation for a single rigid body is given by \cite{park1995lie}:
\begin{equation}
\label{eq:EOM-single}
	J\dot V-\ad_V^*(JV)=F
\end{equation}
where the velocity twist $V \in \se$ and applied external wrench $F \in \se^*$; the inertia tensor $J$ denotes a $6\times 6$ symmetric positive-definite matrix; the dual adjoint map $\ad_V^*$ is identified with the $6\times 6$ matrix $\ad_V^T$. It can be considered as a Euler-Poincar\`e equation defined on \SE \cite{marsden2013introduction}. 

Consider without loss of generality a loopless and branchless (serial) articulated robot with $n$ links. The recursive formulation of dynamics can be derived from \eqref{eq:EOM-single} and summarized by the following global matrix representation \cite{ploen1999coordinate}:
\begin{subequations}
\begin{align} 
		(I-\Gamma)V&=S\dot q + \text{const.}\label{eq:EOM-full-1}\\
   (I-\Gamma)\dot V&=S\ddot q + \ad_{S\dot q}(\Gamma V) + \text{const.}\label{eq:EOM-full-2}\\
        (I-\Gamma)^TF&=J\dot V + \ad_{V}^T(JV) + \text{const.}\label{eq:EOM-full-3}\\
        \tau&=S^TF\label{eq:EOM-full-4}
\end{align}
\label{eq:EOM-full}
\end{subequations}
where for example $V=(V_1^T, \dots, V_n^T)^T$ and $\tau=(\tau_1,\dots,\tau_n)^T$ denote the global velocity and joint torque vectors. The constant terms in \eqref{eq:EOM-full} refer to the velocity and acceleration of the base link or the force applied at the end-effector.
We emphasize that the block lower bi-diagonal matrix $(I-\Gamma)$ (see \cite{ploen1999coordinate}  for more details) represents the forward propagation (from the base to the end-effector) of link velocities $V_i$'s and accelerations $\dot V_i$'s, and dually the block upper bi-diagonal matrix $(I-\Gamma)^T$ represents the backward propagation of external forces $F_i$'s (from the end-effector to the base).

\subsection{Solving bi-diagonal systems using parallel scan operations}
Equation \eqref{eq:EOM-full-1} -- \eqref{eq:EOM-full-4} form a set of linear algebraic equations which can be used to solve both the inverse and forward dynamics problem. On the one hand, the inverse dynamics algorithms to find the applied torques given the motion of robots directly involve the solution of these bi-diagonal systems, and can therefore be accelerated by utilizing state-of-the-art bi-diagonal system solvers. On the other hand, similar bi-diagonal systems also appear in several forward dynamics algorithms. Such bi-diagonal systems can be efficiently solved with the parallel \emph{scan} algorithm as we proposed in~\cite{YWP16a}.

A scan operation takes a binary associative operator $\oplus$ with an identity, and an ordered set $[a_0, a_1, \dots, a_{n-1}]$ of $n$ elements, and returns the vector $[a_0, (a_0 \oplus a_1), \dots, (a_0 \oplus a_1 \oplus \dots \oplus a_{n-1})]$~\cite{blelloch1990prefix}. It can be extended to the 
following linear recursion:
\begin{equation}
 x_i = 
  \begin{cases} 
   c_0 & \text{if } i = 0 \\
   b_{i-1}x_{i-1}+c_{i} & \text{if } 0 < i < n
  \end{cases}
\end{equation}
with the corresponding bi-diagonal system:
\begin{equation}
	\begin{bmatrix}
	   1 &        &        &  \\
	-b_0 & 1      &        &  \\
         & \ddots & \ddots &  \\
	     &        & -b_{n-1} & 1    
\end{bmatrix}\begin{bmatrix}
	x_0\\
    \vdots\\
    x_n
\end{bmatrix}=\begin{bmatrix}
	c_0\\
    \vdots\\
    c_n
\end{bmatrix}
\end{equation}
by setting $a_i$ to be the homogeneous matrix
\begin{equation}
	a_0=\begin{bmatrix}
		1 & c_0\\
        0 & 1
\end{bmatrix}\qquad a_i=\begin{bmatrix}
		b_{i-1} & c_i\\
        0 & 1
\end{bmatrix}, 0< i< n
\end{equation}
and $a_{i-1} \oplus a_i$ is defined as the matrix multiplication $a_ia_{i-1}$.
More further, a bi-diagonal system such as in Equation~\eqref{eq:EOM-full-1}--\eqref{eq:EOM-full-3} can be easily transformed into scan compatible forms \cite{blelloch1990prefix,YWP16a}.
An $O(\log n)$-complexity parallel implementation of scan operation (running on $n$ processors) can be achieved based on the balanced binary tree structure \cite{blelloch1990prefix}.

\subsection{Scan-based forward dynamics algorithms}
The equation of motion in joint coordinates of an articulated robot can be derived by eliminating $V,\ \dot V$ and $F$ from Equation \eqref{eq:EOM-full-1}--\eqref{eq:EOM-full-4} \cite{murray1994mathematical}:
\begin{equation}
M(q)\ddot{q} + C(q, \dot{q})\dot{q} + N(q, \dot{q}) = \tau
\label{eq:eq-motion}
\end{equation}
where $M(q)=S^T(I-\Gamma)^{-T}J(I-\Gamma)^{-1}S$ denotes the joint space inertia (JSI). 
Therefore, one can simply solve the unknown joint acceleration $\ddot q$ by inverting the JSI:
\begin{equation}
\ddot{q} = M(q)^{-1}(\tau - (\underbrace{C(q, \dot{q})\dot{q} + N(q, \dot{q})}_{\ds\tau^{\text{bias}}}))
\label{eq:tau-bias}
\end{equation}
Note that the bias torque $\tau^{\text{bias}}$ in Equation~\eqref{eq:eq-motion} can simply be evaluated using an inverse dynamics solver by setting $\ddot q=0$~\cite{featherstone1984robot}. We shall denote $\tau-\tau^{\text{bias}}$ by $\tau^{\delta}$ for the rest of the paper.

Equation~\eqref{eq:tau-bias} leads to the \emph{joint space inertia inversion algorithms} (JSIIA) (or inertia matrix methods \cite{featherstone1984robot}) for solving the forward dynamics. In particular, we can compute each column of $M(q)$ by solving an inverse dynamics problem by setting the corresponding $\ddot q_i$ to $1$ while setting all $\dot{q}_i$ and all $\ddot q_j (j\neq i)$ to $0$, thereby utilizing the bi-diagonal solver based on scan operations. Inversion of $M(q)$ leads to a time complexity of $O(n^3)$~\cite{siciliano2010robotics,featherstone1984robot}.

The inertia matrix methods are usually outperformed by the $O(n)$ propagation methods \cite{featherstone1984robot}. One of the first examples is the so-called \emph{articulated body inertia algorithm} (ABIA) proposed by Featherstone \cite{featherstone1983calculation}, which hinges on the concept of articulated-body inertia (ABI) that represents the equivalent inertia of a sub-system disassembled from an entire robot~\cite{featherstone1984robot,ploen1999coordinate}. 
It involves the following square factorization of the inverse of $M(q)$~\cite{ploen1999coordinate}:
\begin{equation}
\begin{aligned}
M(q)^{-1}&=[I-S^TY\Pi]^T(S^T\hat{J}S)^{-1}[I-S^TY\Pi] \\
\end{aligned}
\end{equation}
where $\hat J$ is the block diagonal matrix formed by the ABI of consecutive sub-systems, and can be evaluated by a nonlinear recursion \cite{ploen1999coordinate}; $Y = (I-X^T)^{-1}$ and $X$ is a $6n\times6n$ bi-diagonal matrix which implies that we can apply the parallel scan algorithm to efficiently calculate $M^{-1}$ by solving the bi-diagonal system with $(I-X^T)$ or its transpose being the coefficient matrix~\cite{YWP16a}.

\section{Forward Dynamics Algorithms Involving Block Tri-Diagonal Systems}
\label{sec:cfa}
The ABIA is inherently limited by the nonlinear recursion for the derivation of ABI, which can not be accelerated beyond a constant factor by parallel algorithms \cite{hyafil1977complexity,miklovsko1984complexity}.
This limitation is arguably due to the computation of the ABI matrix $\hat J$ based on an alternate elimination of $F_i$'s and $\ddot q_i$'s in Equation \eqref{eq:EOM-full-1}--\eqref{eq:EOM-full-3} \cite{featherstone1984robot}. On the contrary, another strategy used in the so-called \emph{constraint force algorithm} (CFA) \cite{fijany1992parallel} calculates explicitly the global interbody force vector $F$ by eliminating the global acceleration vector $\ddot q$, leading to a Schur complement representation of $M(q)^{-1}$.

\subsection{An elimination strategy with explicit forces}
Using the Lie group notation of \cite{ploen1999coordinate}, we summarize a general elimination approach compatible with CFA as follows. First, we reduce the equations \eqref{eq:EOM-full-1}--\eqref{eq:EOM-full-4} to: 
\begin{subequations}
\begin{align}
	(I-\Gamma)\dot{V} &= S\ddot q \label{eq:qdd}\\
    (I-\Gamma)^TF &=J\dot V \label{eq:F}\\
    \tau^{\delta}&=S^TF \label{eq:tau}
\end{align}
\end{subequations}
by canceling out the bias torque related terms. Next, \eqref{eq:qdd} and \eqref{eq:F} lead to:
\begin{equation}
	\label{eq:cfa-eq0}
	S\ddot q=(I-\Gamma)J^{-1}(I-\Gamma)^TF
\end{equation}
The elimination of $\ddot q$ is then achieved by defining a proper projection matrix $W\in\mathds R^{6n\times 5n}$ such that:
\begin{equation}
\label{eq:cfa-eq1}
	W^TS\ddot q=W^T(I-\Gamma)J^{-1}(I-\Gamma)^TF=0
\end{equation}
for any $\ddot q$ (and therefore $W^TS=0$) and that Equation~\eqref{eq:cfa-eq1} and \eqref{eq:tau} should fully determine $F$, i.e. the co-efficient matrix $H\in\mathds R^{6n\times 6n}$ defined as follows must be non-singular:
\begin{equation}
	\begin{aligned}
		\underbrace{\begin{bmatrix}
			W^T(I-\Gamma)J^{-1}(I-\Gamma)^T\\
	        S^T
		\end{bmatrix}}_{\ds H}F=\tau^{\delta}
    \end{aligned}
\end{equation}

Once $F$ is fully determined, $\ddot q$ can be derived by pre-multiplying $S^T$ on both sides of Equation~\eqref{eq:cfa-eq0}:
\begin{equation}
		S^TS\ddot q=\ddot q=S^T(I-\Gamma)J^{-1}(I-\Gamma)^TF
\end{equation}
where we assume $S^TS = I$ without loss of generality.

\subsection{The Constraint Force Algorithm}

The choice of the projection matrix $W$ is not unique and may eventually lead to different factorization of the inverse of the JSI \cite{fijany2013new}. Here, we shall focus on the original CFA convention where $W=\text{diag}(W_1,\dots,W_n)$ and $W_i\in\mathds R^{6\times 5}$ is a conveniently chosen basis matrix for the constraint force of the $i^{\text{th}}$ joint. 
This results in a decomposition of $F$ into active forces along $S$ and constraint forces $F_c$ along $W$ \cite{fijany1992parallel}:
\begin{equation}
	F=S\tau^{\delta}+WF_c
	\label{eq:force-decomposition-global}
\end{equation}
effectively replacing \eqref{eq:tau}. With this choice of $W$, $H$ can be proved to invertible. Substitute \eqref{eq:force-decomposition-global} into Equation~\eqref{eq:cfa-eq1} leads to:
\begin{equation}
\label{eq:cfa-eq2}
\begin{aligned}
		AF_c&= -B\tau^{\delta}\\
        A&= W^T(I-\Gamma)J^{-1}(I-\Gamma)^TW\\
        B&= W^T(I-\Gamma)J^{-1}(I-\Gamma)^TS
\end{aligned}
\end{equation}
where $A$ must be invertible since the following matrix is the product of non-singular matrices:
\begin{equation}
H\begin{bmatrix}
	W&S
\end{bmatrix}=\begin{bmatrix}
	A & B\\
    0 & I
\end{bmatrix}
\end{equation}
Substitute \eqref{eq:cfa-eq2} back into \eqref{eq:cfa-eq0}:
\begin{equation}
	\begin{aligned}
		\ddot q&=\underbrace{S^T(I-\Gamma)\Ji(I-\Gamma^T)(S}_{\ds C}+W\underbrace{(-A^{-1})B)\tau^{\delta}}_{\ds F_c}	\\
		&=\underbrace{(C+B^T(-A^{-1})B)}_{\ds M(q)^{-1}}\tau^{\delta}
	\end{aligned}
	\label{eq:fijany-final}
\end{equation}
The above equation gives rise to a new factorization of $M(q)^{-1}$ in the form of a Schur complement \cite{fijany1992parallel}. The advantage of this algorithm is that it involves no nonlinear recursion but instead a block tri-diagonal system defined by the block tri-diagonal matrix $A$. As we shall see in the next section, block tri-diagonal systems, like bi-diagonal systems, can also be efficiently solved by parallel algorithms.

\subsection{The Parallel CFA Implementation Steps}
In summary, the CFA first calculates the bias torques using the inverse dynamics algorithm and then subtract them from the input torques. The matrix $M(q)^{-1}$ can be factored as $M(q)^{-1} = C - B^TA^{-1}B$ as shown in~\eqref{eq:fijany-final}. Calculations of $A$, $B$, and $C$ are mutually independent, and only involve constant joint inertia, twists and adjoint transformations, which have already been figured out in the inverse dynamics computation. By exploiting the parallelism of many core GPUs, we can compute $A$, $B$ and $C$ in parallel. After solving the tri-diagonal system~\eqref{eq:cfa-eq2} for constraint forces, the desired joint acceleration vector can be figured out through a final parallel block matrix multiplication $\ddot{q} = C\tau + B^TF_c$. The detailed implementation is illustrated in the Algorithm~\ref{algo:CFA}. In summary, the two building blocks, the bi-diagonal system and the tri-diagonal system, of the unifying scheme for parallel implementation of forward dynamics algorithms are illustrated in Fig.~\ref{fig:big-fig}.
\begin{algorithm}
\DontPrintSemicolon
\caption{$\textsc{CalcFwdDyn\_CFA}([\tau^{\text{in}}_i], [q_i], [\dot{q}_i])$\label{algo:CFA}}
\tcc{Compute the bias torques.}
$[\tau_i^{\text{bias}}] \gets \textsc{CalcInvDyn}([q_i], [\dot{q}_i], 0)$\;
\tcc{Compute the differential torques in parallel.}
$[\tau^{\delta}] \gets \textsc{CalcDiffTorques}([\tau_i^{\text{in}}], [\tau_i^{\text{bias}}])$\;
\tcc{Parallel computations.}
\Begin{
$A \gets \textsc{CalcA}([J_i],[Ad_i],[W_i])$\;
$B \gets \textsc{CalcB}([J_i],[Ad_i],[W_i],[S_i])$\;
$C \gets \textsc{CalcC}([J_i],[Ad_i],[S_i])$\;
$[R_i] \gets -B\tau$
}
\tcc{Solve a block tridiagonal (BTD) system for constraint forces.}
$[F_{i,c}] \gets \textsc{SolvBTDSys}(A, [R_i])$\;
\tcc{Compute joint accelerations.}
$[\ddot{q}_i^\text{out}] \gets \textsc{CalcAcc}([\tau^{\delta}], [F_{i,c}], B, C)$
\end{algorithm}

\externaldocument{related}
\section{Parallel Solver of Block Tridiagonal System}
\label{sec:oee-algo}

Compared to other parts in \eqref{eq:fijany-final} which only involve matrix multiplication, solving the block tri-diagonal system $AF_c = -B\tau$ is more time consuming and thus its performance will have a significant impact on the efficiency of the CFA implementation, especially for systems with a large number of links. Fortunately, the tri-diagonal system can also be solved using parallel techniques, and two well-known solutions are the \emph{cyclic reduction algorithm} (CRA) and the the \emph{recursive doubling algorithm} (RDA)~\cite{Cohen94cyclicreduction, seal2014accelerated}. The parallel RDA is similar to the parallel scan operation that has been used for accelerating the inverse dynamics and parts of the forward dynamics in our previous work~\cite{YWP16a}. However, it needs to assume that all the upper blocks in the tri-diagonal system are nonsingular, which does not hold for the CFA. As a result, we choose to use a variant of CRA called the \emph{odd-even elimination algorithm} (OEE) that does not need such assumption~\cite{levit1989parallel}. The same choice was also made by Fijany in \cite{fijany1995parallel}.

\begin{figure}
\subfigure[]{\includegraphics[width=\columnwidth]{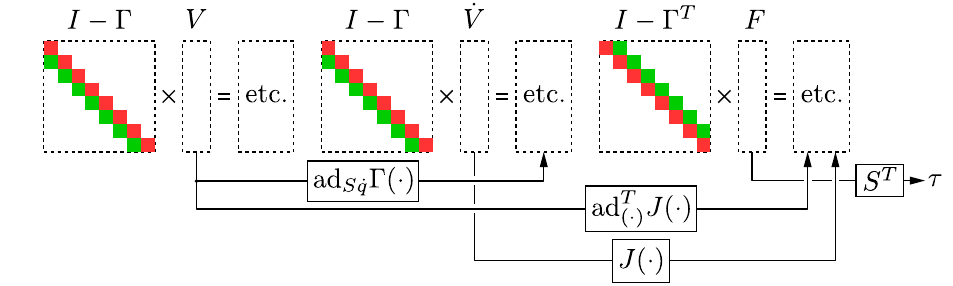}\label{fig:id}}\\
\subfigure[]{\includegraphics[width=\columnwidth]{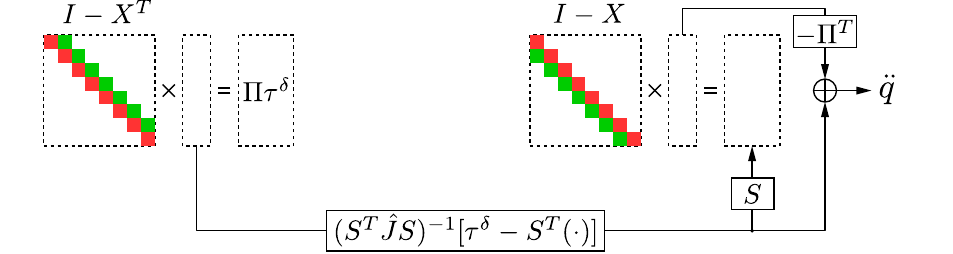}\label{fig:abia}}\\
\subfigure[]{\includegraphics[width=\columnwidth]{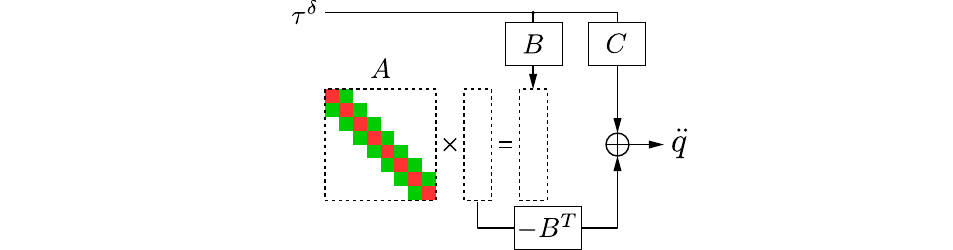}\label{fig:cfa}}\\
\subfigure[]{\includegraphics[width=\columnwidth]{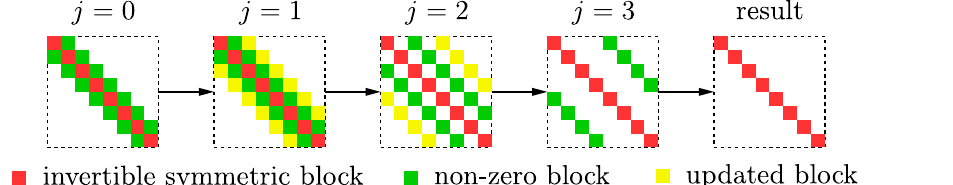}\label{fig:oee-88}}\\
\caption{
(a) Inverse dynamics; (b) ABIA; (c) CFA; (d)
diagonalization of an $8\times8$ block tri-diagonal matrix in the OEE algorithm.}
\label{fig:big-fig}
\end{figure}

Intuitively, given a tri-diagonal system with $n$ diagonal blocks, the OEE algorithm iteratively eliminates the off-diagonal blocks for $\lceil \log_2n \rceil$ times (where $\lceil \  \rceil$ is the ceiling function) until the original tri-diagonal matrix becomes block-diagonal so that every equation in the system can be solved independently. Figure~\ref{fig:oee-88} illustrates the diagonalization process for a linear system with $8$ diagonal blocks, which involves 3 eliminating-updating iterations. In step $j\ (1 \leq j \leq \lceil \log_28 \rceil) $, the non-zero off-diagonal blocks (colored by green) on the $i$-th row are eliminated by subtracting the blocks on the $(i + 2^{j-1})$-th row or the $(i - 2^{j-1})$-th row multiplied by suitable coefficient matrices. The elimination process will update all the diagonal blocks and some off-diagonal blocks (colored by yellow) as shown in Figure~\ref{fig:oee-88}. These updated off-diagonal blocks will be annihilated in the next step in the same manner. Every eliminating or updating calculation only
involve the blocks from the last step and thus can be executed in parallel. 

The $A$ matrix in the CFA is a block tri-diagonal system with the form as follows:
\begin{equation}
A =
\begin{bmatrix}
D_1 & U_1 \\
U_1^T & D_2 & U_2 & 0 \\
\ & \ & \ddots \\
0 & \ & U_{n-2}^T & D_{n-1} & U_{n-1} \\
\ & \ & \ & U_{n-1}^T & D_n
\end{bmatrix}
\label{eq:A_mat}
\end{equation}
This tri-diagonal system has two special properties: First, its diagonal blocks are all symmetric and invertible; second, all the lower off-diagonal blocks are the transpose of the corresponding upper off-diagonal blocks. 
Thus, we only need to update the upper off-diagonal blocks (as shown by the red front blocks in Figure~\ref{fig:oee-88}). Given a linear system of $AF_c = -B\tau$, we can solve it by updating the blocks in the left side and right side in an iterative manner. For the $j$-th step, the update rule is as follows:
\begin{equation}
\begin{aligned}
D_i^{(j)} &= D_i^{(j-1)} - E_i^{(j)}(U_i^{(j-1)})^{T} - K_i^{(j)}U_{i-2^{j-1}}^{(j-1)} \\
U_i^{(j)} &= -E_i^{(j)}U_{i + 2^{j-1}}^{(j-1)}\\
R_i^{(j)} &= R_i^{(j-1)} - E_i^{(j)}X_{i + 2^{j-1}}^{(j-1)} - K_i^{(j)}R_{i - 2^{j-1}}^{(j-1)}
\end{aligned}
\end{equation}
where $D_i^{(j)}$, $U_i^{(j)}$ are the $i$-th diagonal and off-diagonal blocks in the matrix $A$ and $R_i^{(j)}$ is the $i$-th block in $B\tau$. $E_i^{(j)} = U_i^{(j-1)}(D_{i + 2^{j-1}}^{(j-1)})^{-1}$ is the coefficient matrix used to eliminate $U_i^{(j-1)}$ and $K_i^{(j)} = (U_{i-2^{j-1}}^{(j-1)})^T(D_{i - 2^{j-1}}^{(j-1)})^{-1}$ is the coefficient matrix used to eliminate $(U_i^{(j-1)})^T$. By taking the advantage of fact that $D_i$ is symmetric, we can compute the coefficient matrix $E_i^{(j)}$ by solving a linear system  $D_{i + 2^{j-1}}^{(j-1)}(E_i^{(j)})^T=(U_i^{(j-1)})^T$, which is cheaper and more stable than taking the inverse of $D_{i + 2^{j-1}}^{(j-1)}$ directly. The other coefficient matrix $K_i^{(j)}$ can be figured out in the same manner. 
As shown in Figure~\ref{fig:oee-88}, $E_i^{(j)}$ and $D_i^{(j)}$ only need to be calculated for $i = 1\ \text{to}\ n-2^{j-1}$ because the last $2^{j-1}$ upper off-diagonal blocks are removed in the $j$-th step. A more detailed description of this GPU-based parallel OEE algorithm is shown in Algorithm~\ref{algo:OEE}. 

\begin{algorithm}
\DontPrintSemicolon
\caption{$\textsc{OEESolvBTDSys}([D_i^{(0)}], [U_i^{(0)}], [R_i^{(0)}])$\label{algo:OEE}}
\For{$j \leftarrow 1$ \KwTo $\lceil \log_2n \rceil$}{
\tcp{Compute coefficients in parallel for $i = 1$ \KwTo $n-2^{j-1}$.}
\Begin{
$(E_i^{(j)})^T \gets \textsc{SolvSys}([D_{i + 2^{j-1}}^{(j-1)}], [(U_i^{(j-1)})^T])$\;
$(K_i^{(j)})^T \gets \textsc{SolvSys}([D_i^{(j-1)}], [U_i^{(j-1)}])$\;
}
\tcp{Update $D_i$ and $R_i$ in parallel for $i = 1$ \KwTo $n-2^{j-1}$.}
\Begin{
$D_i^{(j)} \gets D_i^{(j-1)} - E_i^{(j)}(U_i^{(j-1)})^T$\;
$D_{i+2^{j-1}}^{(j)} \gets D_{i+2^{j-1}}^{(j)} - K_i^{(j)}U_i^{(j-1)}$\;
$L_i^{(j)} \gets E_i^{(j)}R_{i + 2^{j-1}}^{(j-1)}$\;
$R_{i+2^{j-1}}^{(j)} \gets R_{i+2^{j-1}}^{(j-1)} - K_i^{(j)}R_i^{(j-1)}$\;
$R_i^{(j)} \gets R_i^{(j)} - L_i^{(j)}$\;
}
\tcp{Update $U_i$ in parallel where $i = 1$ \KwTo $n - 2^{j}$.}
$U_i^{(j)} \gets - E_i^{(j)}U_{i + 2^{j-1}}^{(j-1)}$
}
\tcp{Compute all constraint forces in parallel where $j = \lceil \log_2n \rceil$.
}
\Begin{
$[F_{i,c}] \gets \textsc{SolvSys}([D_i^{(j)}], [R_i^{(j)}])$
}
\end{algorithm}

\section{Experiment}
\label{sec:experiment}
In this section, we evaluate the performance of the three parallel dynamics algorithms discussed above by comparing their running time over two experiments. We first investigate the scalability of each algorithm with respect to increasing number of links of a single robot. We next demonstrate the efficacy of each method on a large group of robots with a moderate number of links, which poses as a bottleneck in many applications such as model-based control optimization. The robot configuration parameters such as twists are initialized randomly before the dynamics computation in all our experiments. 

Each experiment will be repeated one thousand times with randomized joint inputs, after which the average computation time is reported. All experiments are conducted on a desktop workstation with an 8-core Genuine Intel i7-6700 CPU and 15.6 GB memory. We implemented our GPU-based parallel dynamics algorithms using CUDA on a Tesla K40c GPU with a 11520 MB video memory and 288GB/sec memory bandwidth.

\subsection{Experiments with Different Link Numbers}
\begin{figure}
\includegraphics[width=\linewidth]{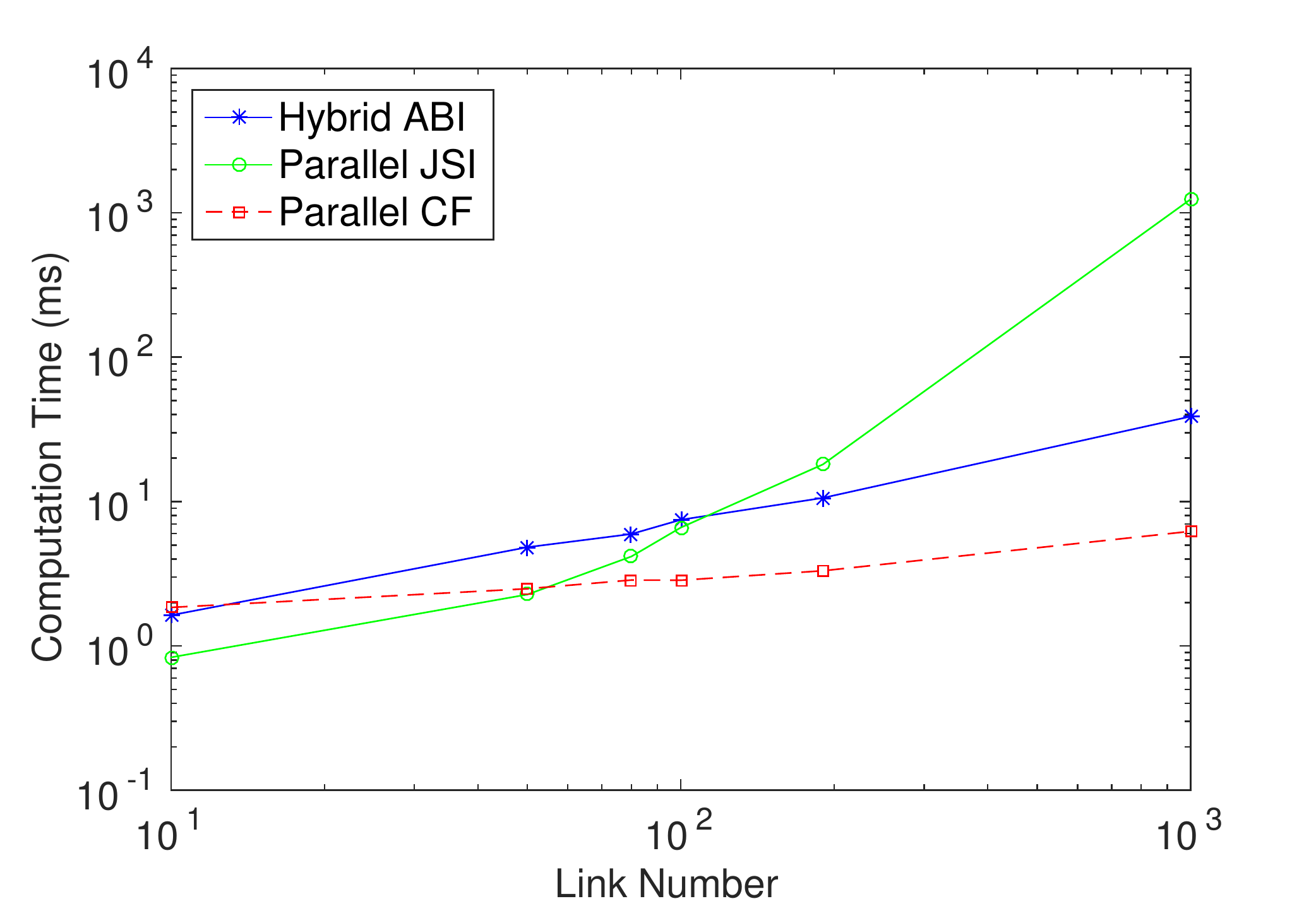}
\caption{Computation time comparison of different parallel forward dynamics algorithms for robots with different number of links}
\label{fig:fd_link_fig}
\end{figure}

We compare the time cost of a single forward dynamics call for robots with a different number of links. We investigate the performance of three different forward dynamics algorithm: the GPU-based parallel JSIIA, the CPU-GPU hybrid ABIA (ABI iteratively computed on CPU; the rest executed on GPUs), and the GPU-based parallel CFA. As illustrated in Figure~\ref{fig:fd_link_fig}, even though the CFA algorithm is computationally more expensive for robots with a small number of links, it eventually outperforms ABIA and JSIIA when the link number becomes substantially large. This is because the CFA introduces some extra operations to achieve the full parallelism. These extra operations result in a relatively high overhead for the situation with a moderate number of links, as explained in~\cite{fijany1995parallel}.
As the system scale increases, such overhead gets mitigated by the high acceleration rate of the other part of the CFA dynamics algorithm. Consequently, the CFA's time cost grows slower as the number of links increases. In comparison, the inversion operation of the joint space inertia matrix (for JSIIA) and the computation of the articulated-body inertia via non-linear recursion (for ABIA) pose as the major bottlenecks for the corresponding parallel forward dynamics when link number is large. 

\subsection{Experiments with Different Group Numbers}
\begin{figure}
\subfigure[Comparison for robots with 10 links]{\label{fig:fd_grp_10links}
\includegraphics[width=\linewidth]{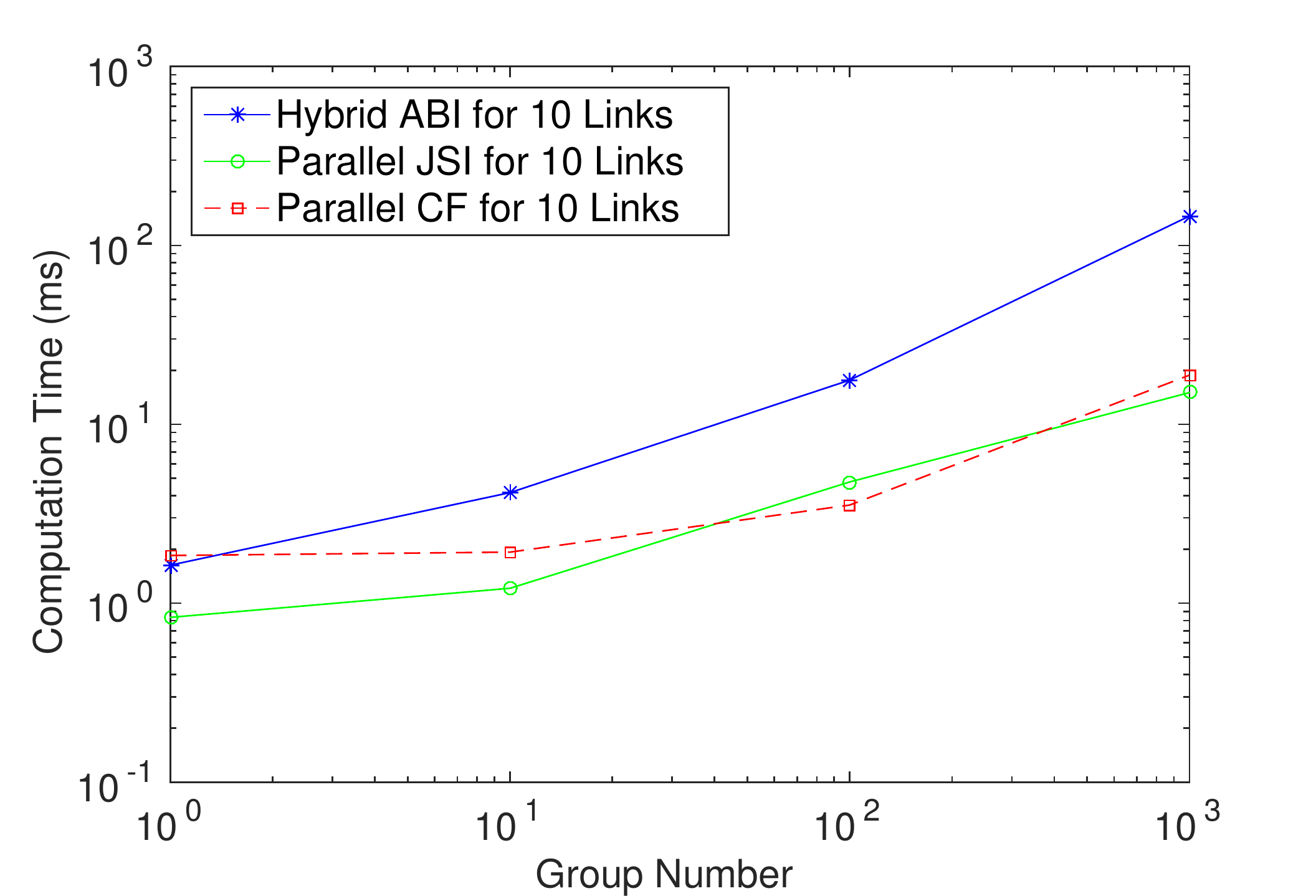}}
\subfigure[Comparison for robots with 200 links]{\label{fig:fd_grp_200links}
\includegraphics[width=\linewidth]{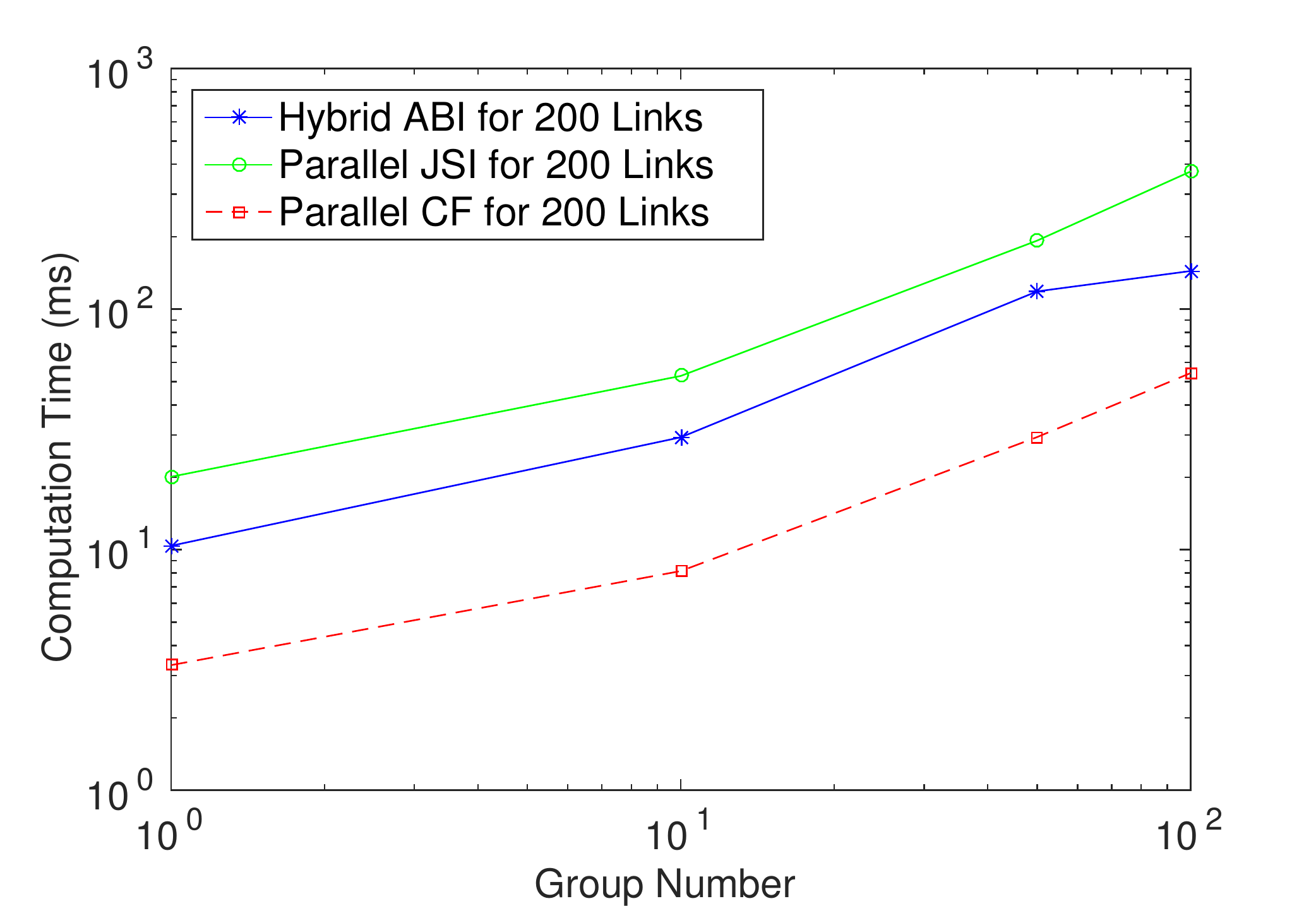}}
\caption{Computation time comparison of different algorithms on many independent forward dynamics calls}
\label{fig:fd_grp}
\end{figure}

We now compare the performances of conducting a large number of independent dynamics computations. Figure~\ref{fig:fd_grp_10links} and Figure~\ref{fig:fd_grp_200links} illustrate the performance of each algorithm for robots with ten links and two hundred links, respectively. For the ten-link robot, the computational efficiency of JSIIA and CFA is similar when the number of groups is larger than one hundred; both are faster than ABIA. However, for the two-hundred-link robot, CFA outperforms the other two approaches, with JSIIA having the worst performance. The performance degradation of JSIIA is due to its matrix inversion operation, whose cost increases cubically with the number of links.

In conclusion, CFA demonstrates the best performance for robots with substantial number of links, while JSIIA is the most suitable for small scale systems. Due to the limited computational efficiency of sequential recursive operations and  overhead for data transfer between CPU and GPU, the performance of ABIA is always dominated by  JSIIA and CFA.  

\section{Conclusion and Future Work}
\label{sec:conclusion}

We have proposed a parallel implementation scheme for various articulated robot forward dynamics algorithms. We address various representative algorithms, such as ABIA, JSIIA and CFA using a unified Lie group notation. We focus on a class of algorithms that admit a direct factorization of the inverse of JSI, and show that different variable elimination strategies lead to different factorization. 

We have identified two common building blocks of a large class (if not all) of forward dynamics algorithms for branchless and loopless articulated robots, namely block bi-diagonal and block tri-diagonal systems. They may be efficiently solved by parallel all-prefix-sum operations (scan) and parallel odd-even elimination (OEE) respectively. We implement the proposed scheme on a Nvidia CUDA GPU platform for the comparative study of JSIIA, ABIA and CFA.

In summary, the CFA provides the best performance when the number of links increases, which conforms to the theoretical analysis presented  in~\cite{fijany1995parallel}. The JSIIA is the fastest algorithm for robots with a moderate number of links, while the hybrid ABIA does not show a high-level competency in all the experiments we conducted.

Our future work includes investigation of general choices of the projection matrix $W$ as in \eqref{eq:cfa-eq1} that lead to factorization of the inverse of JSI into efficiently solvable building blocks. This is eventually an optimization problem for speed and numerical accuracy.

\bibliographystyle{IEEEtran}
\bibliography{ref}

\end{document}